\documentclass[letterpaper, 10 pt, conference]{ieeeconf}  

\IEEEoverridecommandlockouts                              

\let\labelindent\relax

\usepackage{graphics} 
\usepackage{enumitem}
\usepackage{epsfig} 
\usepackage{mathptmx} 
\usepackage{times} 
\usepackage{amsmath} 
\usepackage{amssymb}  
\usepackage{booktabs}
\usepackage{makecell}
\usepackage{algorithm}
\usepackage{algpseudocode}
\usepackage{multirow}
\usepackage{biblatex}
\usepackage{pdfpages}
\usepackage{pifont}
\usepackage{wrapfig}

\DeclareMathOperator*{\argmax}{arg\,max}

\title{\LARGE \bf
 Efficient Q-Learning over Visit Frequency Maps\\ for Multi-agent Exploration of Unknown Environments
}

\author{Xuyang Chen$^{*}$, Ashvin N. Iyer$^{*}$, Zixing Wang and Ahmed H. Qureshi
\thanks{The $^{*}$ indicates equal contribution.}
\thanks{All the authors are with the Department of Computer Science,
        Purdue University, West Lafayette, IN 47906, USA
        {\tt\small [chen4007, aniyer, wang5389, ahqureshi]@purdue.edu}%
}}

\begin{document}

\maketitle
\thispagestyle{empty}
\pagestyle{empty}

\begin{abstract}

The robot exploration task has been widely studied with applications spanning from novel environment mapping to item delivery. For some time-critical tasks, such as rescue catastrophes, the agent is required to explore as efficiently as possible. Recently, Visit Frequency-based map representation achieved great success in such scenarios by discouraging repetitive visits with a frequency-based penalty. However, its relatively large size and single-agent settings hinder its further development. In this context, we propose Integrated Visit Frequency Map, which encodes identical information as Visit Frequency Map with a more compact size, and a visit frequency-based multi-agent information exchange and control scheme that is able to accommodate both representations. Through tests in diverse settings, the results indicate our proposed methods can achieve a comparable level of performance of VFM with lower bandwidth requirements and generalize well to different multi-agent setups including real-world environments.

\end{abstract}

\section{INTRODUCTION}

Exploration is a critical task in robotics with applications ranging from catastrophes rescue to foraging and item delivery~\cite{Liemhetcharat2015MultiRobotID, Lerman2002MathematicalMO, Hamann2006AnAA}. It requires agents to explore the designated unknown environment without any prior knowledge. At the end of a trial of such a task, we expect an accurate map fully or partially covering the task environment. In time-sensitive scenarios such as disaster relief and human rescue, participating agents are tasked to efficiently explore the environment, namely maximize the area explored within a unit time. Previous work~\cite{wang2021spatial} indicates that to achieve such an objective, the robot shall reduce the chance of repetitive exploration, which means revisiting previously seen areas. Visit Frequency Map (VFM)~\cite{wang2021spatial} is a novel state representation encoding the information of the time every location has been visited by the robot, which can reduce the chance of repetitive exploration and avoid the critical no-difference problem of the traditional occupancy map. Incorporating spatial action maps~\cite{wu2020spatial}, a type of action representation with comparatively large and complete action space than the typical steering commands, VFM shows strong performance on efficient novel environment exploration tasks.

Although VFM has solved several vital problems of exploration tasks, there are still challenges to overcome and potential to improve efficiency. Firstly, in some scenarios, a wireless signal is usually weakened by different types of obstruction. VFM, as the state representation that needs to be transmitted very frequently, has a relatively large size as it consists of four different channels of two-dimensional arrays. Thus, we believe a smaller size information encoding format can reduce the latency caused by weak signals and slightly save the required computational power. Secondly, exploration progress can be accelerated by deploying multiple agents to the task environment. As VFM is designed for single-agent setup, we believe an improvement and adaption for multi-agent settings is required to improve the potential of VFM methods. 

To resolve the aforementioned problems, in this study, we propose an Integrated Visit Frequency Map (i-VFM), a state representation encoding identical information as VFM with only half of its size, and a multi-agent control scheme designed for visit frequency class methods (VFM and i-VFM) that exchange necessary data and discourage inter-agent repetitive exploration through frequency information. In conclusion, the contributions of this study can be summarized as follow: 
\begin{itemize}[leftmargin=*]
    \item We propose i-VFM, an informative while compact state representation format for solving robot exploration tasks.
    \item We propose a visit frequency-based multi-agent information exchange and control scheme that scales to an arbitrary number of agents available for exploring unknown environments.
    \item We conduct a series of comprehensive sim and real-world experiments to evaluate the performance and generalization ability of the proposed framework.
\end{itemize}

\section{RELATED WORK}
Environment exploration tasks can be conducted by either active or passive behavior~\cite{fuentes2015visual}. Active methods aim to minimize the uncertainty of environments, while passive methods build environment maps while completing other tasks such as navigation~\cite{gupta2017cognitive}. This section will focus on introducing active exploration methods and drawing their relevance to our proposed framework.

Various types of environment representation have been proposed to plan exploration actions~\cite{xu2017autonomous, parisi2021interesting, Jayaraman_2018_CVPR}. For example,~\cite{xu2017autonomous} proposes to use time-varying tensor fields to plan a smooth path for high-quality environment scanning. Likewise,~\cite{bai2016information} leverage a Bayesian optimization-based predictor to guide the mobile robot to the location estimated to be most informative within its current field of view. 

Neural network-based methods, especially reinforcement learning (RL), proposed a new path for such tasks. Some of them focus on panoramic reconstruction with limited glimpses. Taking~\cite{Jayaraman_2018_CVPR} as an example,  its reward function encourages agents to actively complete panoramic natural scenes with a recurrent neural network-based framework. Apart from that, learning the prior environment arrangement can also enable agents to explore environments effectively. For example,~\cite{ramakrishnan2020occupancy, chaplot2020learning} improve task efficiency by anticipating unseen environment’s structure given the explored area. Our framework also anticipates environmental information. However, instead of predicting explicit structure, our method predicts action-value maps to plan the task. It is worth noting that due to the complexity of the environment and the long duration of the exploration task, planning with a hierarchical framework that can handle global and local goals can improve task efficiency, as~\cite{chaplot2020learning} demonstrated. In addition, some methods build implicit memory structures to discourage repetitive visits, such as~\cite{savinov2018episodic} and~\cite{fang2019scene}.~\cite{savinov2018episodic} uses a curiosity method with a novelty bonus reward formed by episodic memory to encourage exploration, while~\cite{fang2019scene} proposes the Scene Memory Transformer to complete long-horizon tasks including exploration. Compared with them, the VFM and i-VFM directly encode the visit frequency information to the state representation without building extra expensive structures. Besides network structures, the reward function is the key incentive to encourage exploration as highlighted by prior methods \cite{6294131, lopes2012exploration, pathak2017curiosity, stadie2015incentivizing, fu2017ex2}.

In addition to the environment anticipation and past visit encoding advantages, our proposed i-VFM class methods incorporate dense action representation for larger action space and more direct destination assignment. The dense action representation structure has been applied in many areas. For example,~\cite{zeng2022robotic} produces dense probability maps from visual observations that represent pixel-wise affordances for different grasping primitives, and~\cite{wu2020spatial} introduces spatial action maps, a Q-value approximation for dense pixel-wise goal locations reachable by conventional path planning algorithm. Building upon this,~\cite{wang2021spatial} proposes to Visit Frequency Maps to reduce redundant explorations during single-agent foraging tasks.

\section{METHODS}
\begin{figure*}
    \centering
    \includegraphics[scale=0.9,trim={0.2cm 22.3cm 2cm 1.2cm},clip]{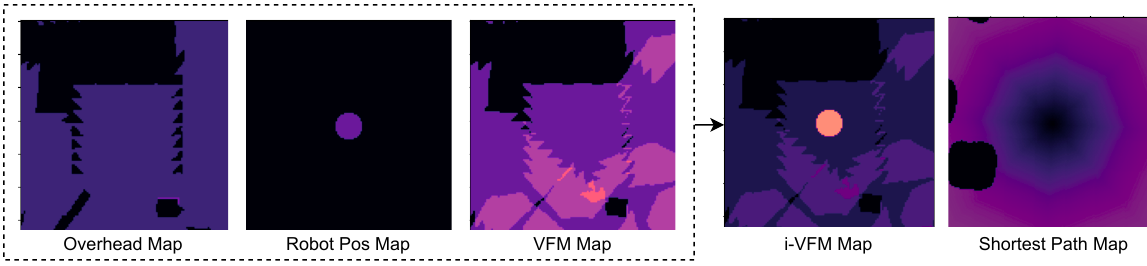}
    \caption{New State Representation channel for i-VFM. The first three channels (overhead map, robot position map, and visit frequency map) are combined into a single i-VFM channel, which is used with the shortest path map as inputs for the policy.}
    \label{fig:ivfm}
\end{figure*}
In this section, we describe our proposed framework for multi-agent exploration in unknown environments. We define a grid-like environment $S \in \mathbb{Z}^2$, along with subsets $S_e \subseteq S$ and $S_u = S - S_e$, which represented explored and unexplored areas respectively. The ranges of $S$ are defined along the x and y axes as $r_x$ and $r_y$ respectively. We define $S(i, j)$ to represent whether the pixel ($i, j$) is explored. 
$$S(i, j) = \begin{cases} 
      0 & (i,j) \in S_u \\
      1 & (i,j) \in S_e 
   \end{cases}
$$
Let $c$ be a metric that determines the percentage of the explored region $S_e$ over the entire region $S$.
$$c(S_e, S) = \frac{\sum_{i}^{r_x}\sum_j^{r_y}S(i,j)}{r_x \cdot r_y}$$
Initially, $S_e = \emptyset$ and $S_u=S$. Our setting assumes a driving robot, equipped with any sensor capable of local observation. The pose of the robot is defined as $p \in \mathbb{P}$, where $\mathbb{P} \in \mathbb{R}^3$ is defined as the x position, the y position, and the yaw angle of the robot. The observation of the sensor is defined as $o \in \mathbb{O}$, where $\mathbb{O} \in \mathbb{R}^n$ is defined as some real output of the sensor. We define a function $F(p, o) \rightarrow S_n$, which takes in a robot pose and observation and outputs a set of pixels $S_n \subseteq S$ that were observed. Let a policy $\pi$ be defined as a function that takes in the current state representation at time $t$ and outputs a singular action to follow. The action is outputted as a single grid coordinate $a_t$. We define a function $G(a_t)\rightarrow p \in \mathbb{P}$, which converts a grid coordinate to a robot pose. A goal robot pose $p_t = G(a_t)$ is found, and the robot moves from its current position to the goal, using any suitable navigation method. As such, we define $S^{t+1}(a_t)$ as a new state in the next iteration after taking action $a_t$. At each iteration, we set $S^{t+1}_e = S^{t}_e \cup S^{t+1}_n$ and $S^{t+1}_u = S - S^{t+1}_e$. Our goal is to develop a policy $\pi$ that outputs actions to maximize the amount of area explored while minimizing the number of actions taken. This is accomplished by maximizing the number of new pixels found at each iteration, leading to this min-max formulation:
\begin{equation}\label{Eq1}
\argmax_{a_t \sim \pi} \min_T \sum_{t=0}^{T-1} c(S_e^{t+1}(a_t) - S_e^t, S)    
\end{equation}
We also introduce multi-agent support for this problem. Each agent will output its own $a_t$, so a set $A_t= \cup_{i=1}^n a^i_t$ can be defined, where $n$ is the number of agents and $a^i_t$ represents the action agent $i$ takes. We now define $S^{t+1}(A_t)$ to be the new state representation after all agents have performed an action. Our new min-max formulation is defined as:

\[
\argmax_{A_t \sim \pi} \min_T \sum_{t=0}^{T-1} c(S_e^{t+1}(A_t) - S_e^t, S)    
\]

This paper extends the original VFM exploration method~\cite{wang2021spatial}, which deploys for single-agent use in standard environments exploration while optimizing for Eq.~\ref{Eq1}. VFM exploration problem is setup as a Reinforcement Learning problem, where a robot is given a state input at each time step and learns to select an appropriate action. VFM's state representation is a stacked set of local maps, each containing information about the same spatial region. At each time step, global maps of information are updated from observations, and local maps are cropped such that it is oriented north and centered around the agent. The state contains 4 channels: 
\begin{itemize}[leftmargin=*]
    \item \textbf{Robot Position Map.} It is a binary mask of the robot's collision radius where 1 indicates obstacle space and 0 indicates obstacle-free space. 
    \item \textbf{Overhead Map.} It contains information about occupancy and segmentation of the explored environment. 
    \item \textbf{Visit Frequency Map.} It stores how many times the robot has observed each pixel.
    \item \textbf{Shortest Path Map.} It gives the distance to each pixel from the robot's given state in the local map.
\end{itemize}


In the original VFM method, this 4-channel state representation is used as the input to a ResNet-based network that outputs a Spatial Action Map,~\cite{wu2020spatial} a dense 2D action space of target locations relative to the robot. Using DQN training, the network learns to estimate the Q-value for each target location, where the highest value serves as the agent's action. However, the learning process for VFMs is inefficient, since they require 4 channel input maps and hence larger networks. Therefore, this paper optimizes for local communication bandwidth to efficiently train neural networks for Q-learning.

\subsection{Integrated Visit Frequency Maps (i-VFM)}
The original VFM method requires a high input size, leading to high bandwidth usage. As a result, environments with unreliable communication are not able to take advantage of traditional VFM exploration. To combat this, we introduce Integrated Visit Frequency Maps (i-VFM): a VFM method that can be utilized in low bandwidth environments.

We observe that an optimal policy is conditioned to minimize the overall time spent in high value regions of the VFM, as the re-exploration penalty is proportionally tied to the local VFM intensity. Thus, we hypothesize that the VFM serves as an implicit artificial repulsive field, where higher values are simply interpreted as less desirable, without the need to differentiate semantic differences between objects. Similarly, we also observe the optimal policy should never output a target that is an obstacle. Aiming for an obstacle is an impossible goal location, so the robot would never feasibly reach its next location. Lastly, the optimal policy should also pick locations further away from the robot so the agent explores faster. Locations close to the robot do not explore many new pixels, as there will be many overlapping pixels between the current and previous scan.

An optimal policy will generally learn to avoid exploring high valued areas from any of the VFM, overhead map, or robot position map. Therefore, we look to combine these 3 channels into 1 singular channel to save bandwidth for the policy, as seen in Fig.~\ref{fig:ivfm}. A simple idea is to combine all of the maps by summing all of there values. However, this quickly becomes problematic, because the occupancy and robot position maps are both binary masks (either 0 or 1), while a VFM value of 1 is relatively small. Scaling the values of the occupancy and robot position map does not solve the issue, because the VFM could eventually make the occupancy and robot position map values insignificant. Any linear combination of the maps cannot provide the desired results as the VFM can contain infinitely high values. Instead, we look to cap the values of the VFM, to combine it with the occupancy and robot position maps. We apply a sigmoid function to the VFM values, to bound them between 0 and a maximum value. Then, we set all pixels of the occupancy and robot position map to contain values higher than the capped sigmoid. The function we used for each pixel in the i-VFM grid is listed below ($O$ - set of occupied pixels, $R$ - set of pixels close to robot):
$$\text{i-VFM}(i, j) = \begin{cases} 
      \lambda \cdot \sigma(\text{VFM}(i, j)) & (i,j) \notin O \cup R  \\
      \lambda + \epsilon & (i,j) \in O \cup R
   \end{cases}
$$
Our function contains 2 parameters, $\lambda$ and $\epsilon$. $\lambda$ represents a scaling factor for the sigmoid function, and $\epsilon$ represents a constant value of how much greater the occupancy and robot position maps values should be compared to the VFM values. The occupancy and robot position map values are set to be $\epsilon$ higher than any possible VFM value, since it better to re-explore an area than run into an obstacle or not move. For our case, we set $\lambda=10$ and $\epsilon=2$. After creating the i-VFM, a new policy is trained with the new map and Shortest Path map, reducing the number of channels used from 4 to 2.

\begin{figure*}
    \centering
    \includegraphics[width=1.0\textwidth]{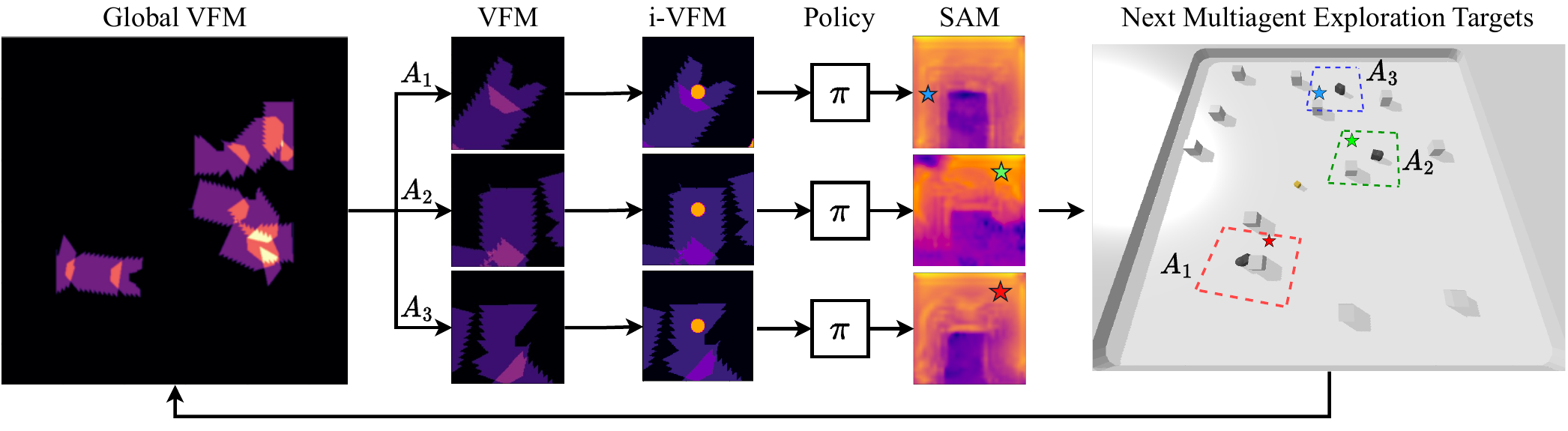}
    \caption{Steps of multi-agent i-VFM: Each robot locally scans its environment. The global maps are updated from observations, and local maps are extracted. The maps are compressed into i-VFM and passed into a policy to select the next robot location, indicated by the stars, for exploration. The heat map range from blue (min) to yellow (max). }
    \label{fig:pipeline}
\end{figure*}
\subsection{Multi-Agent Visit Frequency Maps}

Large areas can be difficult to explore efficiently for a single agent, regardless of how effective the exploration method is. A common solution is to deploy multiple robots in the same area to speed up exploration time. To extend VFM methods to be used in larger areas, we develop a multi-agent VFM exploration.

The simplest way to implement a multi-agent VFM is to run VFM on each individual agent until the target object is found. However, this idea is flawed because there is no collaboration between the robots. If an agent explores a certain region of the map, it is not communicated with the rest of the agents, so they will re-explore the already visited area.

\begin{figure}
    \centering
    \includegraphics[scale=0.78,trim={1.2cm 23.5cm 9.5cm 1.2cm},clip]{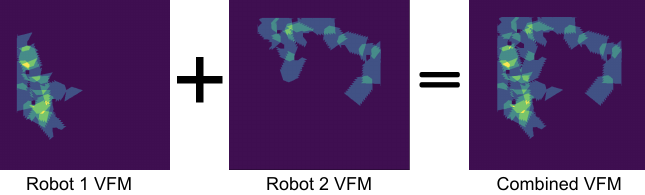}
    \caption{Combined VFM representation for Multi-agent setup. The combined VFM is calculated as the sum of the individual VFM.}
    \label{fig:multiagent}
\end{figure}

We incorporate communication between the agents to provide an efficient multi-agent exploration method. The agents implicitly communicate information with each other through the use of shared global maps. Specifically, the robots exchange information through a shared VFM. Instead of each agent storing its own VFM information, all agents update a single VFM, which is equivalent to the sum of the individual VFMs displayed in Fig.~\ref{fig:multiagent}. A local crop of the shared VFM is taken for each individual agent for the policy. The implicit communication through a shared map is able to effectively prevent agents from re-exploring areas observed by other agents. If a robot explores an area already observed by another robot, the local VFM will show that area as explored, since the global VFM was previously updated at that position. The global overhead map is also shared among all the agents, so agents also have information about obstacles. However, the shortest path map is generated individually for each agent, as each map is relative to the agent's position. Using this method, a multi-agent VFM can be employed without training a new policy, since the representation is built in a way where all explored areas appear as areas explored by the singular agent to the policy. Furthermore, our multi-agent method can be combined with i-VFM because the local map compression works independent of global map sharing, enabling VFM to be used for multi-agent exploration in low bandwidth environments.

Algorithm 1 and Fig.~\ref{fig:pipeline} outlines our multi-agent VFM exploration method. All of the maps are initialized to be empty, as we have no prior information of the environment. At each time step, every robot receives an observation. If any of the robots observed the target object, we end the exploration. Then, the global maps are updated from all observations. After the maps are updated, each robot takes a local crop of the global maps, inputs the local maps into the policy, and receives an action. Each robot follows its corresponding action and keeps exploring until the target object is found.
    \begin{algorithm}[h]
    \begin{algorithmic}[1]  
    \Procedure{multi-agent VFM}{}
    \State $f_{init}(S)$ \Comment{Initialize empty global Map}
    \While {target not found}
    \For {$i \leftarrow 1$ to $n$} \Comment{Updating n agents}
    \State $p$ $\gets$ $F_{p}(robot_i)$ \Comment{Get robot pose}
    \State $o$ $\gets$ $F_{obs}(robot_i, rgbd)$ \Comment{Get RGB-D obs}
    \If {target found}
    \State break
    \EndIf
    \State $S$ $\gets$ $F_{update}(p, o, S)$ \Comment{Update global map}
    \EndFor
    \For{$i \leftarrow 1$ to $n$}
    \State $p$ $\gets$ $F_{p}(robot_i)$
    \State $S_{local}$ $\gets$ $F_{crop}(S, p)$ \Comment{Get local map}
    \State $a_t$ $\gets$ $\pi(S_{local})$  \Comment{Get command from policy}
    \State $F_{move}(a_t)$  \Comment{Execute command}
    \EndFor
    \EndWhile
    \EndProcedure
    \end{algorithmic}
    \caption{Multi-agent VFM-based exploration algorithm}\label{alg:cap}
\end{algorithm}



\section{EXPERIMENTS}
To comprehensively evaluate the performance of our proposed methods, we conducted experiments in both simulated and real environments with single and multi-agent setup. The simulated environment and task setup are introduced in ~\ref{subsec:expstp}. The methods served as baselines are introduced in~\ref{subsec:bl}.~\ref{subsec:metrics} explains the metrics we employ for quantitative quality measurements. The results of our simulated experiment and analysis are reported in~\ref{subsec:resana}. In addition, an experiment to demonstrate our method's real-world task ability is included in~\ref{subsec:real}.

\subsection{Experiment Setup}
\label{subsec:expstp}
\subsubsection{Simulated Environments}
We employ a simulated PyBullet~\cite{coumans2021} scene to serve as our training and evaluation environments. This environment is a square arena bordered by walls that keep the agent within a defined arena. Obstacles are placed within the arena (1) to simulate the unpredictability of navigating unknown regions and (2) to encourage the learning of robust behaviors by reducing the feasibility of straight-line trajectories. The variants of this arena, differing in the overall size and the type of obstacles, are described below. 

\textbf{1X Arena:} The 1x arena is the only environment used for policy training. It is a square 3m x 3m arena with up to 25 square columns, each 0.1m in width, randomly placed throughout. The number of columns was randomly selected for each episode. Agents' performance on this 1x arena serves as a control upon which the effects of other modifications can be compared. \textbf{2X arena:} The 2x arena is sized at 4.2m x 4.2m, or twice the area of the 1x arena, and the number of columns has been increased to 50. The 2x environment measures agents' ability to generalize to differently sized environments. \textbf{Divider:} The divider is a 3m x 3m arena with no columns and is instead bisected by a 0.8m x 0.05m strip that agents cannot cross. The strip is centered on the Y axis, but its X position is randomly initialized. Fig. 4 shows examples of Divider and 2x arenas.


\begin{figure}
\centering
\includegraphics[width=0.23\textwidth]{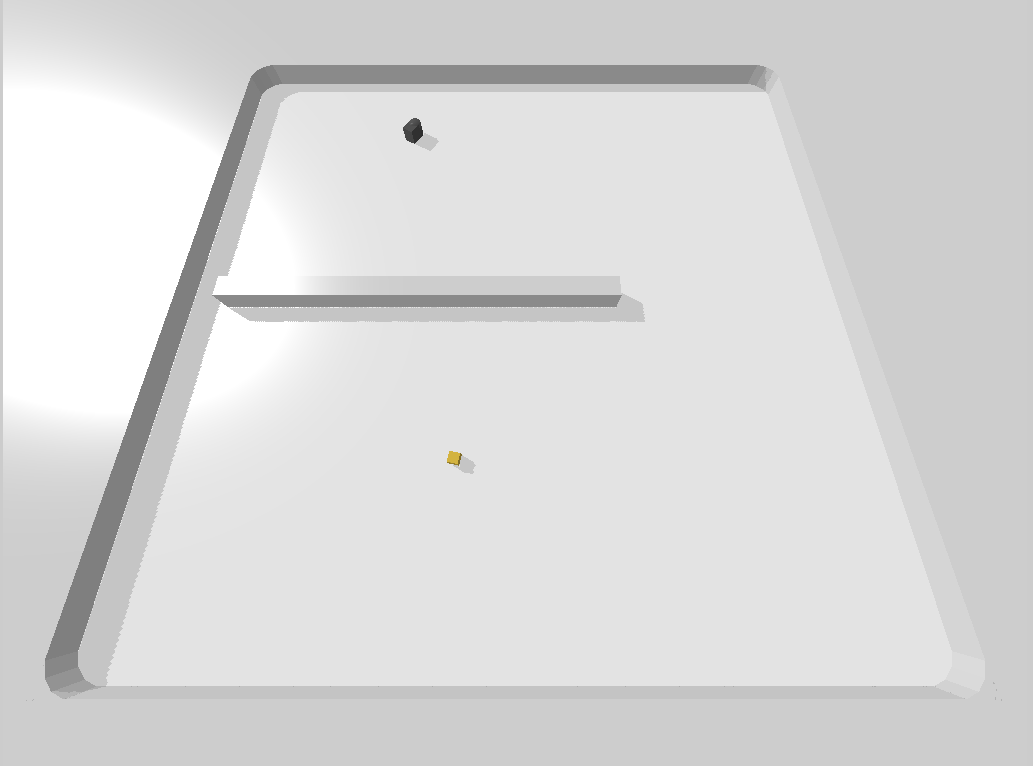}
\includegraphics[width=0.23\textwidth]{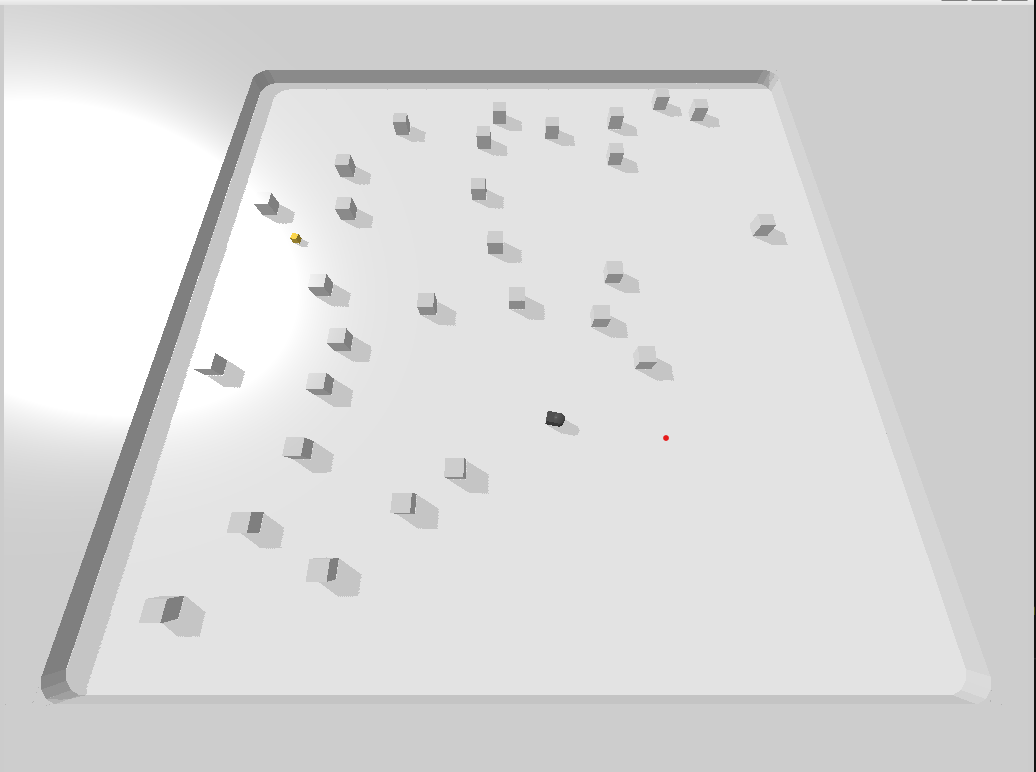}
\caption{The left image shows the divider arena, containing a randomly placed dividing obstacle at the center. The right image shows an example 2x arena with randomly placed columns.}
\end{figure}

\subsubsection{Task Objectives and Setups}
In the above environment setups, the algorithms solve a foraging task where they must explore the arena to locate a randomly placed cube. Upon its discovery, the episode is terminated. For each method, we evaluated 200 episodes. An episode is considered successful if the objective is reached within 5 minutes. Trials exceeding 5 minutes fall into a trend of regressive behavior, and including them would heavily skew the existing metrics. Thus they are excluded from the final calculations. We consider two task setups in those settings:

\textbf{Single Agent Task} experiments are conducted in the 1x arena. In this experiment setup, We aim to test the ability of i-VFM and other baselines and evaluate the trade-off between performance by comparing the result of VFM and i-VFM.

\textbf{Multi-Agent task} examines the feasibility of a multi-agent exploration setup, where agents jointly participate in a cooperative foraging task where the episode is terminated when any agent discovers the cube. All policies used in this task are conditioned purely on the single agent scenario but share a common global VFM and occupancy map. This means of implicit communication serves as the only medium for cooperation in these trials. 

Both single and multi-agent trials begin with the agents initialized at random non-colliding points in the arena. This uniform allocation leads to the possibility of disadvantageous initial positions, where agents may be clustered in a small region and be unable to subdivide the arena. Both positive and disadvantageous situations are accounted for in the final metrics presented for evaluation. Furthermore, each agent has a virtual RGBD camera with a 60-degree field-of-view that allows it to differentiate columns, cubes, and walls, and fully compute the state representation. 

\subsection{Baselines}
\label{subsec:bl}
\textbf{Random:} The Random baseline uniformly samples goal points from the same action space as i-VFM. This baseline was not evaluated in larger environments due to high time consumption caused by the extremely high failure rate. In addition, since the random policy does not take into account other agents of the VFM, we feel that the multi-agent results were unnecessary. 

\textbf{Frontier Exploration:} We evaluate a Frontier Exploration (FE) policy as benchmark against the performance of our trained agents. The FE policy is evaluated on the single agent foraging task and produces the same metrics as the other trials. Frontier Exploration~\cite{613851} is a grid-based algorithm that involves detecting the boundaries between explored and unexplored regions and moving towards them. Frontiers are naturally updated to reflect newly explored regions. The openness of our simulation environment often leads to the generation of a single closed frontier. While the original implementation used centroids as the target position, such would lead to deadlocks for our environment as the centroid of a closed frontier is often far from its border. Taking care to retain as much of the base algorithm as possible, we replaced centroids with medians instead. Due to its low efficiency, this baseline was not evaluated on larger environments in the interest of time.

\subsection{Evaluation Metrics}
\label{subsec:metrics}
The following \textit{metrics} are used to access the performance of all methods.

\textbf{Repetitive Exploration Rate (RER)} The RER measures the efficiency of the policy during exploration. RER is computed as the average number of visits to a unit area:
$$\text{RER} = \frac{\sum^{\text{size(S)}}{S - J}}{\text{size(S)}},$$
where $size(S)$ is the size of the VFM, $J$ is a matrix of all ones with the same dimension as $S$.

\textbf{Path Efficiency (PE)}. The PE measures the efficiency of an average path. It is computed as the average area explored per distance traveled. $$ \text{PE} = \frac{\text{size}(E)}{\text{length}(D)}, $$
where E is the total area explored, and length(D) represents the cumulative distance the agent traveled before the termination condition.

\textbf{Overlap Ratio}: The Overlap Ratio is a measure of multi-agent exploration efficiency. It is the ratio of area explored by more than one agent per the total current area explored.
\[o(S_e, S) = \frac{\prod_{i}^A {S^i > 0}}{\sum^{\text{size}(S)} {S > 0}},\]
where $S^i$ is the individual visit frequency map for agent $i$, $A$ is the total number of agents, and $size(S)$ is the region bounding the global visit frequency map.

\textbf{Coverage}: The ratio of area explored collectively by all agents over the total explorable area. 
\[v(S_e, S) = \frac{\sum^{\text{size}(S)}S(i,j)} {E},\]
where $size(S)$ is the region bounding the global visit frequency map and $E$ is the size of the total explorable area.

\textbf{Bandwidth (BW)}: 
During our experiments, we simulate the bandwidth usage of agents by tracking the size of the data exchanged between a central server and each agent. This metric sums the total transmitted bytes for all agents and is displayed in MiB. 

The total bandwidth includes the size of transmitted state representations. VFM uses a 4-channel state representation of size $M^{4 \cdot size(S)}$ whereas the i-VFM policy uses a compressed state of size $M^{2 \cdot size(S)}$. Both representations are expressed in \texttt{uint16}. Agents send newly observed obstacles $\Delta O$ and VFM updates $\Delta S$ to the central server as they explore. As the non-zero portions of these updates are usually sparse in relation to the global map size, agents only transmit the minimum bounding rectangles \texttt{MBR($\Delta O$)} and \texttt{MBR($\Delta S$)}. Since the Minimum Bounding Rectangles sent in step 2 lack fixed positions,  additional $(x, y, \theta)$ poses of type \texttt{float16} give the server context for where the data is positioned on the global map.

\textbf{Step Count}: The average number of steps taken per episode summed over all agents.

\textbf{Not Found:} The number of episodes in which the termination condition was not reached within 5 minutes.

\begin{table*}[t]
    \centering
    \caption{Evaluation Performance in 1x Sized Arena}
    \begin{tabular}{ c | c | c c c c c c c } 
      \toprule
         Policy & Agents & RER $\downarrow$ & PE $\uparrow$ & Steps $\downarrow$ & Overlap $\downarrow$ & Bandwidth $\downarrow$ & Coverage $\uparrow$ & Not Found $\downarrow$\\
        \midrule
         Random Action & One & 3.083 $\pm$ 2.551 & 2972 $\pm$ 1924 & 145.5 $\pm$ 161.4 & N/A & N/A & 0.490 $\pm$ 0.327 & 3 / 200 \\
         
        \midrule
        Frontier Exploration & One & 1.957 $\pm$ 2.074 & 12469 $\pm$ 12792 & 44.7 $\pm$ 50.4 & N/A & $>$ 36 & 0.487 $\pm$ 0.306 & 8 / 200 \\
        
        \midrule
            \multirow{3}{*}{VFM} 
                & One   & 0.488 $\pm$ 0.597 & 6252 $\pm$ 2259 & 33.9 $\pm$ 41.6 & N/A & 4.6 $\pm$ 5.6 & 0.501 $\pm$ 0.325 & 0 / 200 \\
                & Two   & 0.457 $\pm$ 0.426 & 6033 $\pm$ 2157 & 32.7 $\pm$ 32.2 & 0.1 $\pm$ 0.1 & 4.4 $\pm$ 4.3 & 0.503 $\pm$ 0.306 & 0 / 200 \\
                & Four  & 0.387 $\pm$ 0.306 & 5930 $\pm$ 2481 & 30.8 $\pm$ 24.4 & 0.1 $\pm$ 0.1 & 4.2 $\pm$ 3.3 & 0.478 $\pm$ 0.305 & 0 / 200 \\
            \\[-6 pt] \cline{1-9} \\[-6 pt]
            \multirow{3}{*}{i-VFM} 
                & One   & 0.622 $\pm$ 0.639 & 5827 $\pm$ 2157 & 40.0 $\pm$ 43.7 & N/A & 2.9 $\pm$ 3.1 & 0.525 $\pm$ 0.322 & 0 / 200 \\
                & Two   & 0.612 $\pm$ 0.609 & 5607 $\pm$ 2301 & 39.9 $\pm$ 39.5 & 0.1 $\pm$ 0.1 & 2.9 $\pm$ 2.8 & 0.508 $\pm$ 0.309 & 2 / 200 \\
                & Four  & 0.484 $\pm$ 0.480 & 6086 $\pm$ 2468 & 36.3 $\pm$ 34.5 & 0.1 $\pm$ 0.1 & 2.6 $\pm$ 2.5 & 0.467 $\pm$ 0.309 & 0 / 200 \\

      \bottomrule
    \end{tabular}
    \label{tab:1x}
\end{table*}

\begin{table*}[t]
    \centering
    \caption{Evaluation Performance In Divider and 2x Sized Arenas}
    \begin{tabular}{ c | c | c | c c c c c c c } 
      \toprule
        Trial & Policy & Agents & RER $\downarrow$ & PE $\uparrow$ & Steps $\downarrow$ & Overlap $\downarrow$ & Bandwidth $\downarrow$ & Coverage $\uparrow$ & Not Found $\downarrow$ \\
        \midrule
        \multirow{6}{*}{Divider}
            & \multirow{3}{*}{VFM} 
                  & One & 0.492 $\pm$ 0.517 & 5914 $\pm$ 2250 & 31.8 $\pm$ 34.3 & N/A & 4.3 $\pm$ 4.6 & 0.480 $\pm$ 0.308 & 6 / 200 \\
                & & Two & 0.477 $\pm$ 0.473 & 5601 $\pm$ 2353 & 32.3 $\pm$ 33.0 & 0.1 $\pm$ 0.1 & 4.4 $\pm$ 4.4 & 0.466 $\pm$ 0.316 & 0 / 200 \\
                & & Four & 0.502 $\pm$ 0.496 & 5574 $\pm$ 2626 & 35.5 $\pm$ 35.6 & 0.2 $\pm$ 0.1 & 4.8 $\pm$ 4.7 & 0.473 $\pm$ 0.319 & 0 / 200 \\
            \\[-6 pt] \cline{2-10} \\[-6 pt]
            & \multirow{3}{*}{i-VFM} 
                  & One & 0.584 $\pm$ 0.424 & 5524 $\pm$ 2305 & 33.6 $\pm$ 30.0 & N/A & 2.4 $\pm$ 2.1 & 0.460 $\pm$ 0.302 & 5 / 200 \\
                & & Two & 0.577 $\pm$ 0.529 & 5529 $\pm$ 2376 & 32.7 $\pm$ 30.8 & 0.1 $\pm$ 0.1 & 2.4 $\pm$ 2.2 & 0.443 $\pm$ 0.293 & 1 / 200 \\
                & & Four & 0.602 $\pm$ 0.651 & 5664 $\pm$ 2519 & 39.3 $\pm$ 41.8 & 0.2 $\pm$ 0.2 & 2.8 $\pm$ 3.0 & 0.472 $\pm$ 0.289 & 0 / 200 \\
        \midrule
        \multirow{10}{*}{2x Arena}  
            & \multirow{5}{*}{VFM} 
              & One   & 0.484 $\pm$ 0.471 & 6136 $\pm$ 1961 & 59.3 $\pm$ 62.9 & N/A & 8.1 $\pm$ 8.5 & 0.477 $\pm$ 0.315 & 1 / 200 \\
            & & Two   & 0.507 $\pm$ 0.522 & 6206 $\pm$ 1909 & 62.4 $\pm$ 67.9 & 0.1 $\pm$ 0.1 & 8.5 $\pm$ 9.1 & 0.473 $\pm$ 0.319 & 1 / 200 \\
            & & Three & 0.422 $\pm$ 0.401 & 5995 $\pm$ 2223 & 54.4 $\pm$ 54.2 & 0.1 $\pm$ 0.1 & 7.4 $\pm$ 7.3 & 0.444 $\pm$ 0.297 & 1 / 200 \\
            & & Four  & 0.419 $\pm$ 0.343 & 6230 $\pm$ 2078 & 54.8 $\pm$ 49.6 & 0.1 $\pm$ 0.1 & 7.5 $\pm$ 6.7 & 0.453 $\pm$ 0.306 & 1 / 200 \\
            & & Five  & 0.375 $\pm$ 0.303 & 6867 $\pm$ 1804 & 49.9 $\pm$ 44.9 & 0.1 $\pm$ 0.1 & 6.8 $\pm$ 6.1 & 0.440 $\pm$ 0.290 & 3 / 200 \\
            \\[-6 pt] \cline{2-10} \\[-6 pt]
            & \multirow{5}{*}{i-VFM} 
              & One   & 0.575 $\pm$ 0.427 & 5957 $\pm$ 1696 & 66.5 $\pm$ 59.5 & N/A & 4.8 $\pm$ 4.3 & 0.485 $\pm$ 0.287 & 1 / 200 \\
            & & Two   & 0.550 $\pm$ 0.409 & 5735 $\pm$ 2084 & 66.8 $\pm$ 58.6 & 0.1 $\pm$ 0.1 & 4.8 $\pm$ 4.2 & 0.477 $\pm$ 0.310 & 1 / 200 \\
            & & Three & 0.494 $\pm$ 0.370 & 6072 $\pm$ 2062 & 61.8 $\pm$ 52.8 & 0.1 $\pm$ 0.1 & 4.5 $\pm$ 3.8 & 0.452 $\pm$ 0.293 & 0 / 200 \\
            & & Four  & 0.491 $\pm$ 0.366 & 6083 $\pm$ 1992 & 61.9 $\pm$ 51.1 & 0.1 $\pm$ 0.1 & 4.5 $\pm$ 3.7 & 0.461 $\pm$ 0.290 & 3 / 200 \\
            & & Five  & 0.420 $\pm$ 0.354 & 6421 $\pm$ 2188 & 56.1 $\pm$ 51.5 & 0.1 $\pm$ 0.1 & 4.1 $\pm$ 3.7 & 0.406 $\pm$ 0.292 & 3 / 200 \\
      \bottomrule
    \end{tabular}
    \label{tab:2x}
\end{table*}

\setlength{\textfloatsep}{0.1cm}
\begin{figure}
    \centering
    \includegraphics[width=\linewidth]{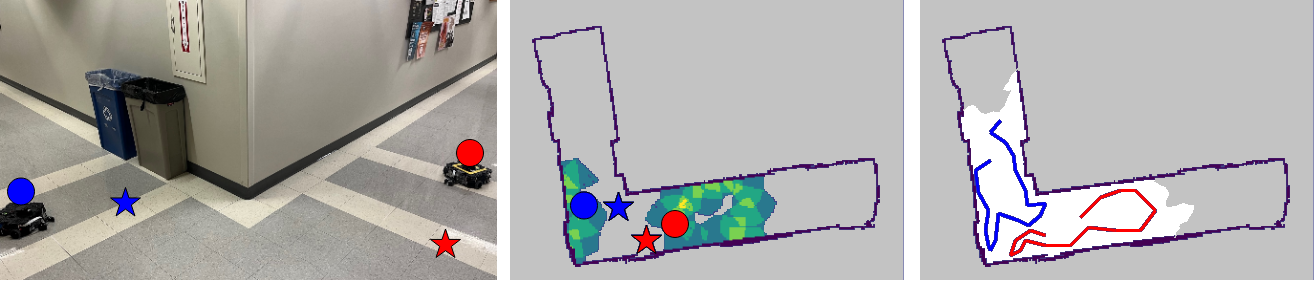}
    \caption{The left image shows the real-world setup with the robot locations (circle) and next goals (star). The middle image shows the VFM overlay with the ground truth map, the current robot positions, and goals. The right image shows both the full robot trajectories after 50\% of the area is explored.}
    \label{fig:real}
\end{figure}

\subsection{Results and Analysis}
\label{subsec:resana}


\subsubsection{1x Arena Experiments}
We report the results of all trials evaluated on the 1x arena in Table~\ref{tab:1x}. According to the results, we have the following analysis: 

\textbf{Performance of i-VFM compared to VFM}
Comparing the metrics between VFM and i-VFM, i-VFM consistently demonstrates slightly worse RER and PE metrics. However, the reduction in state size consistently saves about 35\% bandwidth overall. Single agent scenarios favor the use of VFM, but i-VFM is clearly preferable multi-agent configurations where bandwidth is in demand or communication is unreliable.

The combined two channel state representation sacrifices a small amount of semantic information that is otherwise available to the VFM policy. We had speculated that the Visit Frequency Map and the obstacles were viewed by the policy as `undesirable' regions without differentiation. We believe this assumption is generally correct, as the i-VFM efficiency is close to that of VFM. 


The RER, PE, and Step Count metrics appear to be tightly correlated. Agents with a better RER tend to have a higher rate of exploration and thus higher PE. Agents with a faster rate of exploration tend to discover the target cube in less steps, which reflects in the step count metric.

\textbf{Performance of VFM and i-VFM compared to Baselines}
The Random Action and Frontier Exploration policies show significant inefficiencies and higher step costs. From the large disparity in RER, we can infer that VFM and i-VFM policies are able to reduce re-explorations in a large part by using information in the Visit Frequency Map during action selection. Unexpectedly, the Frontier benchmark outperforms the VFM/i-VFM trials in the PE metric twofold. This is because the Frontier Exploration algorithm can select target positions from the whole arena, while the VFM/i-VFM policy is limited to a fixed action space. As expected, Frontier Exploration requires much higher bandwidth to transmit the complete global map and has a higher failure rate.  

\subsubsection{2x and Divider Generalization Experiments}
In this series of evaluation, we test the generalization ability to large size environment and multi-agent setup of our proposed framework. The results of trials conducted in 2x and Divider arenas are reported in Table~\ref{tab:2x}. The analysis of the related results are:

\textbf{Performance of i-VFM and VFM in Divider and Large Environments}

VFM and i-VFM exploration performance in the 2x arena is comparable to that of the 1x arena. Agents often move in a spiral pattern, starting from the boundaries of the arena and working their way in. This way, the VFM forms a ring that guides the agent as it successively explores more of the edges of the arena. This strategy is adaptable to arenas of arbitrary size, but we note that the step count increases proportionally to the area explored.

Although our policies perform well in the 2x arena, it is possible they may not generalize to environments with different types of obstacles. The divider arena introduces a central obstacle that is much larger than seen in the training environment. Moreover, unlike the columns, agents cannot easily move around the divider. However, both VFM and i-VFM only show minor performance degradation compared to the 1x arena. We conclude that VFM and i-VFM are also able to generalize to unseen obstacles that are difficult to navigate around. 

\textbf{Multi-Agent Generalization Experiment}
Overall, individual agents perform better as the swarm size is increased, requiring less steps and exploring new regions more efficiently. We speculate that when agents share the arena with others, they each individually divide and conquer the space inside the arena, independently exploring smaller regions. A single agent is less efficient because it may need to cross over previously seen areas. However, for the multi-agent case, agents use the VFM to avoid areas already being explored by others. Since we do not employ any inter-agent communication, scaling by agents only increases the bandwidth linearly.

The results illustrate that the total step count and bandwidth among all agents are proportional to the explorable area but remarkably \textit{invariant} to the number of agents. Our state formulation, combined with the limited horizon of DQN training, restricts the network from understanding the greater structure of the environment. Instead, it learns emergent exploration behaviors that are agnostic from the configuration of global features.

\subsection{Real World Generalization Experiment}
\label{subsec:real}
A real-world experiment is conducted with the policy trained in a simulated environment to evaluate the sim-to-real transfer ability of our proposed framework. We deployed 2 Turtlebot3 robots in the corridor of a large building as Fig.~\ref{fig:real} shows. These robots were equipped with an RPLIDAR-A1 and ran ROS. We were interested in how exploration performance was influenced by sensor noise, odometry drift, and the length of the corridor which deviated from the squareness of our simulated arenas. We obtained a ground truth map of the corridor beforehand with the ROS gmapping package. We initialized both robots at random configurations and allowed the policy to run until it explored 50\% of the environment.

According to the results reported in Table~\ref{tb:real}, our multi-agent control scheme generalizes well to the real world without retraining or reconfiguration. While the performance is slightly worse than the simulated trials, it is expected as the policy must account for sensor and localization noise in a real-world setting. Additionally, since our environment is a narrow corridor with defined ends, the agents may reach a corner and be forced to backtrack in order to find fresh unexplored areas. The agents perform backtracking successfully but at the penalty of slightly higher RER and lower PE.

\begin{table}[h]
\centering
    \caption{Evaluation Performance in Real World Environment}
    \begin{tabular}{ c c c c } 
      \toprule
       RER $\downarrow$ & PE$\uparrow$ & Steps $\downarrow$ & Coverage $\uparrow$ \\ 
      \midrule
      0.626 $\pm$ 0.135 & 4533 $\pm$ 410 & 42.5 $\pm$ 26.2 & 0.514 $\pm$ 0.008 \\
      \bottomrule
    \end{tabular}
    \label{tb:real}
\end{table}
\section{CONCLUSION}
In this paper, we proposed i-VFM and a frequency-based multi-agent information exchange and control scheme. According to the result of our evaluation, we believe the high encoding density of i-VFM and scalability of our multi-agent framework enable more efficient exploration under constrained environments. In the future, possible directions to improve our proposed framework are (1) integrating the path representation into i-VFM to further compress its size and (2) lifting the application scenario from 2D to 3D for challenging environment modeling and exploration.




\printbibliography

\end{document}